\documentclass[letterpaper]{article} 
\usepackage[draft]{aaai2026}  
\usepackage{times}  
\usepackage{helvet}  
\usepackage{courier}  
\usepackage[hyphens]{url}  
\usepackage{graphicx} 
\usepackage{natbib}  
\usepackage{caption} 
\usepackage{algorithm}
\usepackage{algorithmic}
\usepackage{booktabs}
\usepackage{multirow}
\usepackage{graphicx}
\usepackage{booktabs}
\usepackage{amsmath} 
\usepackage{amssymb} 
\usepackage{subcaption}

\usepackage{newfloat}
\usepackage{listings}
\DeclareCaptionStyle{ruled}{labelfont=normalfont,labelsep=colon,strut=off} 
\floatstyle{ruled}
\newfloat{listing}{tb}{lst}{}
\floatname{listing}{Listing}
\title{MultiMedEdit: A Scenario-Aware Benchmark for Evaluating Knowledge \\ Editing in Medical VQA}

\author{
    Shengtao Wen\textsuperscript{\rm 1}\equalcontrib,
    Haodong Chen\textsuperscript{\rm 1}\equalcontrib, 
     Yadong Wang\textsuperscript{\rm 1}, 
    Zhongying Pan\textsuperscript{\rm 2},\\ 
    Xiang Chen\textsuperscript{\rm 1}\thanks{Corresponding author.}, 
    Yu Tian\textsuperscript{\rm 3}, 
    Bo Qian\textsuperscript{\rm 1},
    Dong Liang\textsuperscript{\rm 1}, 
    Sheng-Jun Huang\textsuperscript{\rm 1},  
}
\affiliations{
    \textsuperscript{\rm 1}
MIIT Key Laboratory of Pattern Analysis and Machine Intelligence,\\
College of Computer Science and Technology,\\
Nanjing University of Aeronautics and Astronautics \\
\textsuperscript{\rm 2}
Huaneng Information Technology Co., Ltd. 
\textsuperscript{\rm 3}
Tsinghua University \\

  \{xiang\_chen\}@nuaa.edu.cn 
}

\usepackage{bibentry}

\begin{document}

\maketitle

\begin{abstract}
Knowledge editing (KE) provides a scalable approach for updating factual knowledge in large language models without full retraining. While previous studies have demonstrated effectiveness in general domains and medical QA tasks, little attention has been paid to KE in multimodal medical scenarios. Unlike text-only settings, medical KE demands integrating updated knowledge with visual reasoning to support safe and interpretable clinical decisions. To address this gap, we propose \textbf{MultiMedEdit}, the first benchmark tailored to evaluating KE in clinical multimodal tasks. Our framework spans both \textit{understanding} and \textit{reasoning} task types, defines a three-dimensional metric suite (reliability, generality, and locality), and supports cross-paradigm comparisons across general and domain-specific models. We conduct extensive experiments under single-editing and lifelong-editing settings. Results suggest that current methods struggle with generalization and long-tail reasoning, particularly in complex clinical workflows. We further present an efficiency analysis (e.g., edit latency, memory footprint), revealing practical trade-offs in real-world deployment across KE paradigms. Overall, MultiMedEdit not only reveals the limitations of current approaches but also provides a solid foundation for developing clinically robust knowledge editing techniques in the future.
\end{abstract}

\section{Introduction}

\begin{figure}[t]

    \centering
   \includegraphics[width=\linewidth]{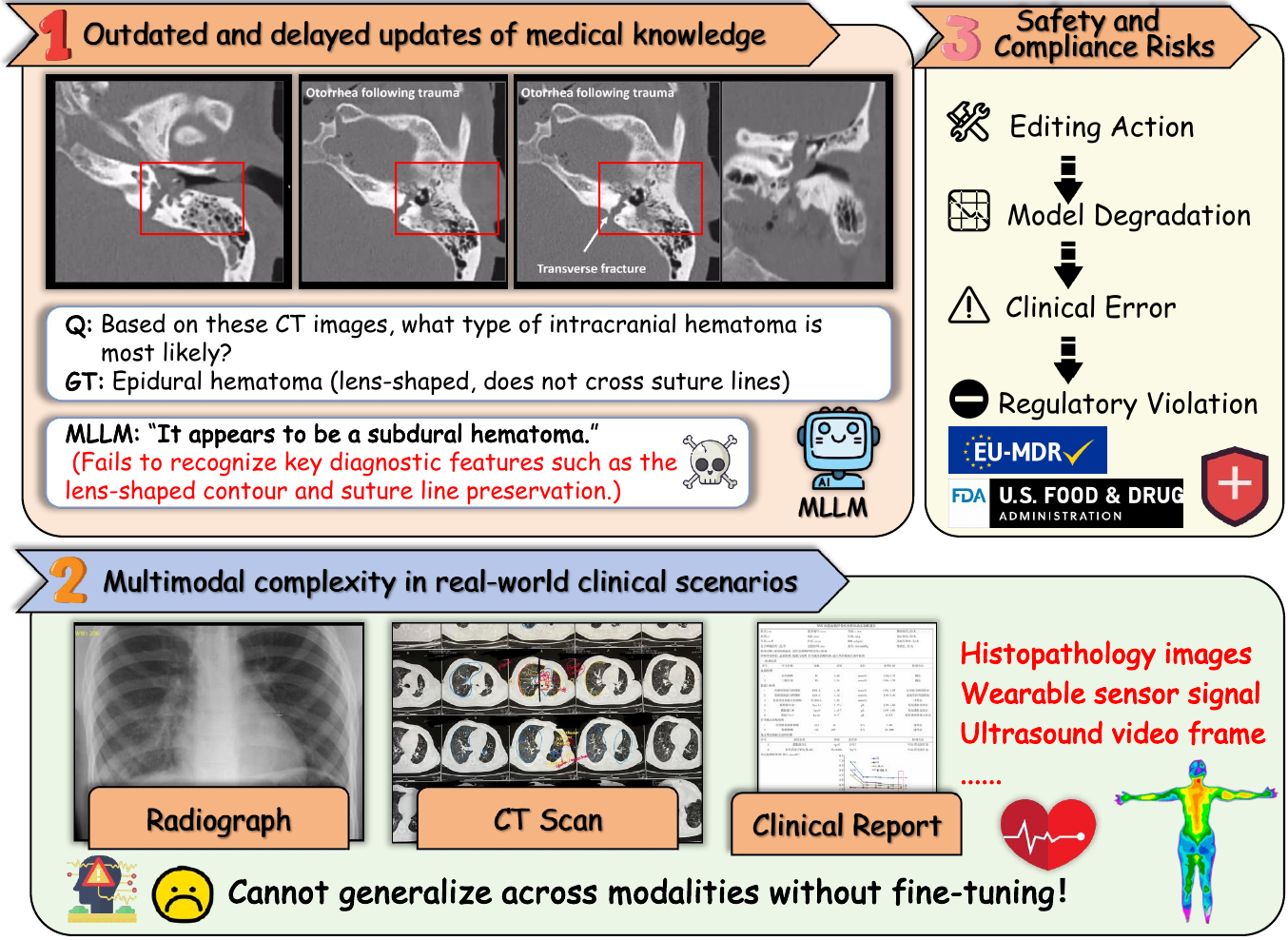}
    \caption{Key challenges faced by general-purpose MLLMs in clinical applications: (1) outdated medical knowledge can lead to inaccurate or unsafe outputs; (2) high diversity in data modalities and task types demands robust multimodal understanding; (3) safety-critical requirements necessitate traceable and reliable model behavior.}
    \label{fig:3Questions}
\end{figure}

Multimodal Large Language Models (MLLMs) have witnessed rapid advancements, with breakthroughs in visual-language alignment and scalable model architectures enabling strong performance across a range of general tasks. However, applying these models effectively in high-stakes domains such as medicine remains an open challenge. Despite initial success in tasks like medical image interpretation, clinical question answering, and decision support, solely relying on model scaling or domain adaptation remains far from sufficient. As illustrated in Figure~\ref{fig:3Questions}, general-purpose MLLMs still face structural limitations in clinical settings—including outdated knowledge~\citep{wu2025driftmedqa,kim2025medicalhallucination}, high heterogeneity across modalities~\citep{unihd,dai2025climb}, and strict demands for safety, interpretability, and  compliance~\citep{wang2024safetyChallenges,han2024medSafetyBench,workum2025safeLLM}.

Real-world medical practice evolves continuously with the approval of new therapies, revisions of treatment guidelines, and discoveries from ongoing clinical trials. In such a dynamic and rapidly changing landscape, deploying frozen MLLMs trained on static corpora creates a widening gap between model behavior and clinical reality. This highlights a crucial but underexplored challenge: how can we efficiently and safely inject new medical knowledge into existing MLLMs without compromising prior capabilities? Traditional fine-tuning offers limited solutions to these challenges~\citep{luo2023empiricalCF}. It typically requires full-network optimization, incurs high computational costs, and involves long iteration cycles. Furthermore, it is prone to catastrophic forgetting, where new knowledge overwrites existing competencies and severely degrades performance on previously learned tasks~\citep{kalajdzievski2024scalingCF,li2024revisitingCF}.


Knowledge Editing (KE) has emerged as a more surgical and efficient alternative to traditional fine-tuning. Instead of retraining the entire model, KE introduces localized updates—via selective parameter modification or external memory injection—allowing models to incorporate new facts rapidly while preserving global behavior~\cite{mitchell2021fastedit,yao2025cake}. Its efficiency, low interference, and auditability make it especially suited for safety-critical domains like healthcare~\cite{youssef2025tracing}.

Despite recent advances in KE for general-purpose language models, its applicability to clinical domains remains largely underexplored. Existing benchmarks, such as MedEditBench~\citep{xu2024medkebench}, focus primarily on textual edits and often overlook challenges fundamental to real-world healthcare, including multimodal fusion, clinical reasoning, among others. To truly assess and enhance the potential of models in real-world healthcare settings, the field urgently requires a new, multimodally-centered benchmark.

To address this need, we present MultiMedEdit, a benchmark specifically designed for evaluating knowledge editing in clinical multimodal contexts. Unlike existing knowledge-editing benchmarks such as MedEditBench, which are limited to textual QA and static fact correction. MultiMedEdit is the first to support \textit{multimodal} and \textit{VQA-based} evaluation in medical settings. This introduces unique challenges: models must not only incorporate new knowledge, but also interpret complex image-text inputs, localize lesions, reason across time, and adapt to diverse clinical modalities. These requirements significantly raise the bar for reliable and safe knowledge editing in real-world clinical scenarios.
The benchmark encompasses the following components:
(1) \textbf{Dual-Axis Task Design}: Tasks are structured along two dimensions, including task type (\textit{understanding} vs. \textit{reasoning}) and input modality (text plus single-frame or multi-frame images), covering the full spectrum from visual recognition to multimodal diagnostic inference.
(2) \textbf{Three-Dimensional Evaluation Metrics}: We propose a comprehensive framework consisting of reliability (accuracy on edited targets), generality (robustness to semantic variations), and locality (preservation of unrelated outputs), enabling systematic analysis of both effectiveness and side effects.
(3) \textbf{Cross-Paradigm Method Comparison}: We evaluate four representative KE paradigms, namely Prompt, LoRA~\citep{zhang2023adalora}, GRACE \citep{wang2024grace}, and WISE~\cite{wang2024wise}, under both single and lifelong editing settings across general-purpose MLLMs and domain-specific medical models.

Extensive experiments yield three key findings. First, existing KE methods underperform on complex long-tail reasoning tasks in medical contexts. Second, lifelong editing introduces order dependence and catastrophic forgetting, reducing model stability. Third, most methods are limited to short-text or atomic fact edits and cannot support the depth and contextual richness required in realistic clinical scenarios. In general, our contributions are threefold:
\begin{itemize}
    \item We propose the first comprehensive benchmark specifically designed for multimodal medical knowledge editing, targeting critical challenges in clinical AI evaluation.
    \item A unified three-dimensional evaluation framework is established to quantify editing effectiveness, generalization, and unintended side effects in medical settings.
    \item Extensive empirical analysis highlights the critical limitations of current KE methods in handling complex clinical reasoning, thereby offering a foundation for future research and method development.
\end{itemize}

We envision MultiMedEdit as a crucial stepping stone toward clinically reliable knowledge updating protocols in future medical foundation models.

\section{Related Work}

\subsection{Knowledge Editing for LLMs}

Knowledge Editing seeks to enable accurate and semantically coherent responses from large language models (LLMs) by selectively updating internal knowledge representations without resorting to full model retraining \citep{yao2023editing}. As LLMs scale and are increasingly deployed in real-world applications, ensuring the timeliness and factual consistency of their embedded knowledge becomes essential, particularly in high-stakes fields like healthcare.

Previous studies have shown that most LLMs struggle to adapt to time-sensitive updates unless explicitly augmented or edited \citep{li2024kne,wu2024updating,dhingra2022timeaware}. To address this, research has proposed various knowledge editing paradigms \citep{comprehensive}, broadly categorized as: (1) \textbf{External Retrieval-Based Approaches}, which avoid modifying model weights by retrieving relevant facts from external memory or tools \citep{ike,serac,li2025mindbridge,lte}; (2) \textbf{Latent State Injection}, which integrates new knowledge by altering internal representations \citep{grace,melo,lora,qlora,adalora}; and (3) \textbf{Internal Structure Editing}, which directly edits model parameters to encode persistent knowledge updates \citep{rome,memit,mend,alphaedit,o-edit,feng2025geoedit}.

\subsection{Toward Knowledge-Adaptive MLLMs in Medicine}

Multimodal large language models (MLLMs) such as Flamingo \citep{alayrac2022flamingo}, PaLI \citep{chen2022pali}, and GPT-4V \citep{yang2023dawn} have significantly advanced vision-language tasks, including VQA, image captioning, and multimodal instruction following. Extending this progress to the medical domain, models like HuatuoGPT-Vision \citep{chen2024huatuogpt}, LLaVA-Med \citep{li2023llavamed}, and Med-Flamingo \citep{boger2023medflamingo} demonstrate that domain-specific fine-tuning on radiological images and clinical reports improves grounding and interpretability. XrayGPT \citep{zhang2023xraygpt} and ChatRad \citep{huang2024chatrad} further explore diagnostic reasoning and radiology-focused VQA, indicating early promise in real-world clinical tasks.

However, deploying MLLMs in clinical settings remains challenging. Medical applications require models to maintain up-to-date medical knowledge, yet most existing systems are trained statically and lack mechanisms for post-deployment updates. This can lead to hallucinations and outdated recommendations \citep{yan2024probmed,sepehri2024mediconfusion}. Knowledge editing offers a solution by enabling targeted model updates without full retraining. Early efforts like MedLaSA \citep{xu2024editingmedicalllms} adopt adapter-based strategies to improve factual accuracy in medical LLMs. Nonetheless, most prior work remains unimodal and lacks support for editing visual knowledge. Furthermore, current benchmarks primarily evaluate general-domain or text-only edits \citep{he2024mmllms,hu2024omnimedvqa}, leaving an important gap in evaluating multimodal, domain-specific model adaptation. To bridge this gap, we introduce the first benchmark dataset for multimodal knowledge editing in medical MLLMs.

\section{Benchmark Construction}
\begin{figure*}[t]

    \centering
   \includegraphics[scale=0.53]{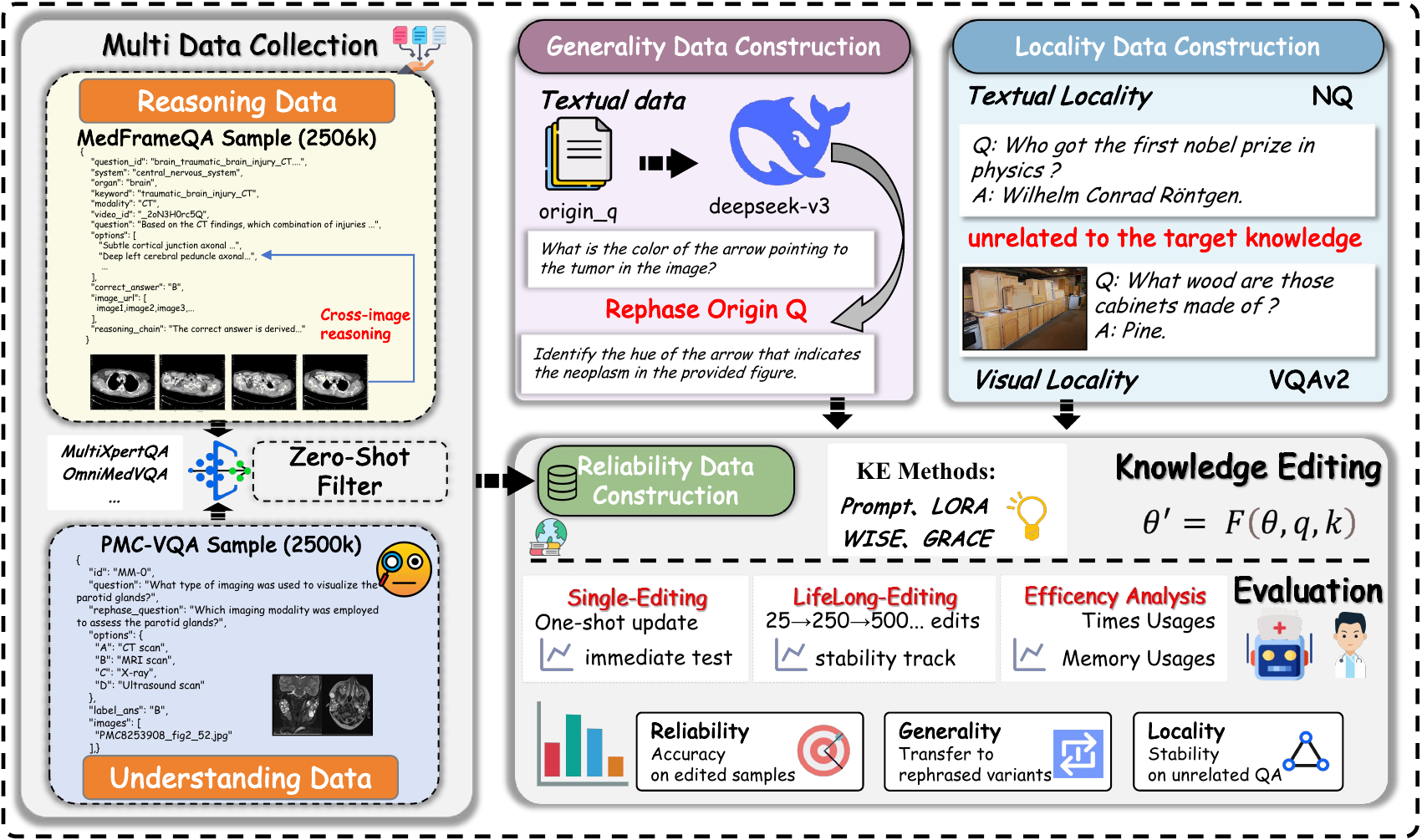}
    \caption{Overview of the MultiMedEdit pipeline. Evaluation subsets (reliability, generality, locality) are built via zero-shot filtering and external sources (DeepSeek-V3, VQAv2, NQ). Four editing methods (Prompt, LoRA, GRACE, WISE) are tested on two general MLLMs—LLaVA-OneVision and Qwen2-VL—and one medical MLLM, HuatuoGPT, across multiple criteria.}
    \label{fig:Overview}
\end{figure*}

Inspired by recent knowledge editing applications in unimodal settings, the constructed MultiMedEdit dataset is composed entirely of visual question answering data. The construction process is illustrated in Figure \ref{fig:Overview}.

\subsection{Preliminaries}

Knowledge editing aims to adjust the behavior of a base model $f_\theta$ (where $\theta$ denotes the model parameters) with respect to a edit descriptor $(x_e, y_e)$, while preserving the model’s performance on other samples. The ultimate goal is to obtain an edited model, denoted as $f_{\theta_e}$~\cite{yao2023editing}.

The base model $f_\theta$ is defined as a function $f: X \to Y$ that maps an input $x$ to a predicted output $y$. Given an edit input $x_e$ and target label $y_e$ such that $f_\theta(x_e) = y_e$, the edited model is expected to satisfy $f_{\theta_e}(x_e) = y_e$.

Knowledge editing typically affects a set of inputs closely associated with the edit example, referred to as the \textit{editing scope}. A successful edit should modify the model’s behavior within the scope, while keeping predictions unchanged for out-of-scope inputs. Formally:
\begin{equation}
f_{\theta_e}(x) = 
\left\{
\begin{array}{ll}
y_e, & \text{if } x \in I(x_e, y_e) \\
f_\theta(x), & \text{if } x \in O(x_e, y_e)
\end{array}
\right.
\end{equation}

Here, $I(x_e, y_e)$ denotes in-scope inputs, typically including $x_e$ and its equivalence neighborhood $N(x_e, y_e)$; whereas $O(x_e, y_e)$ refers to inputs unrelated to the edit descriptor.

\begin{figure}[htp]

    \centering
   \includegraphics[scale=0.25]{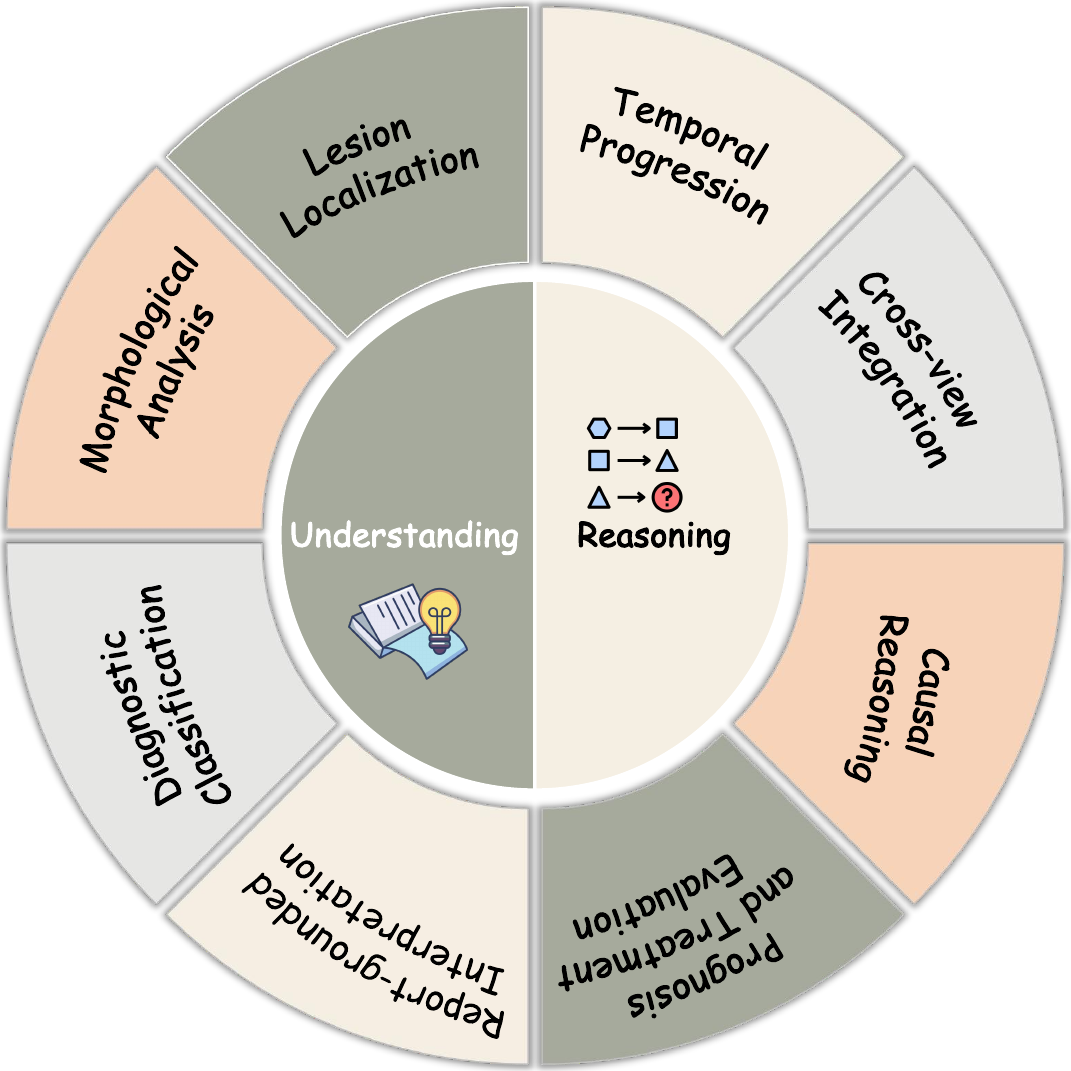}
    \caption{The statistics of scenario types for MultiMedEdit, encompassing two principal categories of clinical tasks: \textit{understanding} and \textit{reasoning}.}
    \label{fig:scenarioTypes}
\end{figure}

\subsection{Design Principle}
To systematically evaluate medical multimodal knowledge-editing methods, we adopt a two-tier experimental taxonomy: the \textit{Understanding} tier requires models to integrate medical images with clinical narratives to deliver a coherent explanation of the patient’s condition, whereas the \textit{Reasoning} tier demands cross-view and temporal inference over multi-frame studies to support complex clinical decisions. This hierarchical design assesses model competence from basic visual recognition to full multimodal diagnostic reasoning. The benchmark encompasses two input modalities: \textit{single-frame} images and \textit{multi-frame} images. Multiple metrics are designed to evaluate knowledge editing methods, such as reliability, generality and locality.

\paragraph{Scenario Type.}
The \textit{Understanding} scenario requires models to fuse one or a few medical images with accompanying clinical narratives—such as chief complaints, radiological findings, and medical history—to produce a semantically consistent explanation covers lesion location, characteristics, staging, and potential management strategies. 
The \textit{Reasoning} scenario represents a more demanding tier: models perform cross-view and temporal inference over multi-frame or multi-view studies, tracking lesion dynamics and jointly leveraging imaging, textual, and temporal cues to support disease-course analysis, treatment-response evaluation, and prognostic prediction. 
Relative to the \textit{Understanding} tier, the \textit{Reasoning} tier places greater emphasis on temporal modelling, cross-view alignment, and causal reasoning. Statistics for each data type are summarised in Table~\ref{tab:table1}. The distribution of scenario types is provided in Figure~\ref{fig:scenarioTypes}.

\paragraph{Data Type.}
The \textit{Single-frame} modality comprises a single static CT or MRI slice, or an ultrasound frame, accompanied by a concise clinical narrative.  It targets foundational visual perception and local diagnostic tasks, such as lesion localisation and morphological assessment.  
The \textit{Multi-frame} modality consists of time-series or multi-view images acquired from the same anatomical region, together with aligned textual descriptions, and specifically is intended to probe a model’s ability in temporal modelling, cross-view fusion, and dynamic lesion analysis.

\paragraph{Metrics.}

\textit{Reliability} denotes the post-editing hit rate on the target samples, reflecting whether the intended knowledge has been correctly injected. \textit{Generality} measures the proportion of correct responses under semantically equivalent paraphrases, gauging the transferability of the edit. \textit{Locality} quantifies the extent to which predictions on unrelated tasks or samples remain unchanged after editing, thereby assessing the edit’s side-effect footprint.

\subsection{Data Collection}
To construct a representative, challenging, and evaluation–friendly benchmark for multimodal medical knowledge editing, we perform a systematic filtering and hierarchical restructuring of publicly available resources. The resulting dataset comprises multiple high-quality subsets tailored to two core tasks: \textit{Understanding} and \textit{Reasoning}.

\paragraph{Data Sources.}
We construct our comprehensive medical multimodal knowledge editing benchmark based on three high-quality public datasets: MedFrameVQA~\citep{liu2024medframeqa} for reasoning-oriented tasks, PMC-VQA~\citep{he2023pmcvqa} for understanding-oriented tasks, and selected samples from both MedXpertQA~\citep{cheng2024medxpertqa} and OmniMedVQA~\citep{huang2024omnimedvqa} to enhance task diversity and overall knowledge coverage.

To ensure the benchmark’s difficulty and diagnostic value, we employ a \textit{zero-shot filtering} strategy. We use medical MLLM \texttt{Radiology-Infer-Mini} to perform inference over all raw samples and retain only those for which the model fails to produce correct answers. This results in two challenging subsets, \textit{MultiMedEdit\textsubscript{U}} and \textit{MultiMedEdit\textsubscript{R}}, designed to focus evaluation on model deficiencies and increase the signal-to-noise ratio.

\paragraph{Construction of the Reliability Subset.}
The filtered samples from \textit{MultiMedEdit\textsubscript{U}} and \textit{MultiMedEdit\textsubscript{R}} together constitute the Reliability subset, which is specifically used to assess whether injected knowledge is correctly reflected in the model’s updated responses.

\begin{table}[ht]

    \centering    

    \begin{tabular}{lccc}
        \toprule[1.5pt]
        \multicolumn{1}{c}{\textbf{Subset}} & 
        \multicolumn{1}{c}{\textbf{Understanding}} & 
        \multicolumn{1}{c}{\textbf{Reasoning}} & 
        \multicolumn{1}{c}{\textbf{Total}} \\
        \midrule
        \multicolumn{1}{c}{\textbf{\textit{Training}}} & 
        \multicolumn{1}{c}{3161} & 
        \multicolumn{1}{c}{3257} & 
        \multicolumn{1}{c}{6418} \\
        \multicolumn{1}{c}{\textbf{\textit{Testing}}} & 
        \multicolumn{1}{c}{1355} & 
        \multicolumn{1}{c}{1396} & 
        \multicolumn{1}{c}{2751} \\
        \bottomrule[1.5pt]
    \end{tabular}
    \caption{Number of samples by task type and modality in MultiMedEdit}
    \label{tab:table1}
\end{table}

\paragraph{Construction of the Generality Subset.}
To evaluate a model’s generalization ability under linguistic variation, we construct the Generality-Text subset. Using the DeepSeek-v3 model, we generate modified versions of the original questions that preserve semantic meaning and ground-truth answers, resulting in semantically equivalent question pairs with various surface forms. Details on prompt templates and rewriting heuristics are provided in the Appendix D.

\paragraph{Construction of the Locality Subset.}
To assess the potential side effects of knowledge editing on seemingly unrelated tasks, we construct two locality-focused evaluation subsets. The first, Locality-Text, is randomly sampled from the Natural Questions dataset~\citep{kwiatkowski2019natural} to quantify whether the edited model still maintains its performance on general-domain language understanding tasks. The second, Locality-Modality, is based on VQAv2~\citep{antol2015vqa} and includes open-domain natural image question-answer pairs to measure whether the model successfully retains general visual reasoning capabilities in non-medical contexts after medical knowledge editing.

\paragraph{Quality Control.} 
To enhance the dataset's validity and robustness, we have implemented a multi-stage quality control pipeline. For consistency and ambiguity, three graduate-level annotators have reviewed a random 20\% sample of paraphrased and generalized question-answer pairs to assess semantic equivalence and clarity, and have resolved disagreements by majority vote. To validate this review process, we have measured inter-annotator consistency across 200 random samples, which has revealed a disagreement rate of less than 7\%. These discrepancies have been settled through majority voting and adjudicated revision.

\begin{table*}[ht]

    \begin{center}
        \centering
        \small
        \setlength{\tabcolsep}{0.8mm}
        \begin{tabular}{cc *{12}{l}}
            \toprule[1.5pt]
            \multirow{2}{*}{\textbf{Methods}} & \multirow{2}{*}{\textbf{Task Type}} 
            & \multicolumn{4}{c}{\textbf{LLaVA-Onevision}} 
            & \multicolumn{4}{c}{\textbf{QWen2-VL}} 
            & \multicolumn{4}{c}{\textbf{HuaTuoGPT-7B}} \\
            \cmidrule(lr){3-6}
            \cmidrule(lr){7-10}
            \cmidrule(lr){11-14}
            & & Rel. & Gen. & T-Loc. & M-Loc. 
              & Rel. & Gen. & T-Loc. & M-Loc. 
              & Rel. & Gen. & T-Loc. & M-Loc. \\
            \midrule[0.5pt]

            \multirow{2}{*}{Prompt}
            & \textit{Understanding} & 0.9749* & 0.9565* & 0.7729 & 0.7986 & 0.8871* & 0.5380* & 0.7874 & 0.7367 & 0.0170 & 0.0236 & 0.8151 & 0.8099 \\
            & \textit{Reasoning}     & 0.9413† & 0.9234† & 0.8055 & 0.8440 & 0.3897 & 0.4936† & 0.7865 & 0.7522 & 0.0953 & 0.2693† & 0.8076 & 0.8040 \\
            \midrule[0.5pt]

            \multirow{2}{*}{WISE}
            & \textit{Understanding} & 0.5058 & 0.5122 & 1.0000* & 1.0000* & 0.5262 & 0.4593 & 1.0000* & 1.0000* & 0.8915* & 0.8155* & 1.0000* & 1.0000* \\
            & \textit{Reasoning}     & 0.3997 & 0.3818 & 1.0000† & 1.0000† & 0.4599 & 0.4248 & 1.0000† & 1.0000† & 0.7550† & 0.6956† & 1.0000† & 1.0000† \\
            \midrule[0.5pt]

            \multirow{2}{*}{LoRA}
            & \textit{Understanding} & 0.4546 & 0.4458 & 0.9987 & 0.9989 & 0.1852 & 0.1756 & 0.9990 & 0.9980 & 0.9033* & 0.8871* & 0.9748 & 0.9531 \\
            & \textit{Reasoning}     & 0.3102 & 0.3209 & 0.9993 & 0.9981 & 0.0874 & 0.0788 & 0.9995 & 0.9951 & 0.8152† & 0.7787† & 0.9792† & 0.9669 \\
            \midrule[0.5pt]

            \multirow{2}{*}{GRACE}
            & \textit{Understanding} & 0.9173 & 0.4295 & 1.0000* & 1.0000* & 0.8827 & 0.1616 & 1.0000* & 1.0000* & 0.5417 & 0.0007 & 1.0000* & 1.0000* \\
            & \textit{Reasoning}     & 0.2872 & 0.2973 & 1.0000† & 1.0000† & 0.9964† & 0.0695 & 1.0000† & 1.0000† & 0.5294 & 0.0215 & 1.0000† & 1.0000† \\
            \bottomrule[1.5pt]
        \end{tabular}
        
    \end{center}
    \caption{
    Evaluation results of four editing methods under single edit. Asterisks (*) indicate the highest value per column in the \textit{Understanding} scenario, and daggers (†) mark the highest in \textit{Reasoning}.
    }
    \label{tab:table2}

\end{table*}

\subsection{Editing Method Selection}
To broadly compare intrinsic knowledge editing paradigms, we select representative methods from three major categories. LoRA represents the fine-tuning paradigm, using low-rank adapters to efficiently update a subset of weights. WISE exemplifies memory-enhanced editing via addressable external memory to overwrite outdated knowledge. GRACE represents parameter-injection approaches by introducing new parameter matrices to encode knowledge with minimal disruption to original weights. We also include a lightweight prompt-based method that edits via input prompt without altering model parameters. Retrieval-augmented methods are excluded, as they rely on external sources rather than modifying internal representations, and thus fall outside the scope of intrinsic editing~\citep{song2024blackbox,shi2024retrieval,chen2024lifelong,zheng2023incontext}.

\section{Experiments}
\subsection{Experimental settings}
We conduct experiments using two state-of-the-art open-source MLLMs: \texttt{LLaVA-OneVision} and \texttt{Qwen2-VL}. Both models share the Qwen2-7B language backbone but differ significantly in their visual encoding architectures and multimodal alignment strategies. Building upon these models, we evaluate four representative knowledge editing paradigms: Prompt, LoRA, GRACE and WISE. To eliminate the confounding effects introduced by visual modality variations, all editing methods are applied exclusively to the language components, with the visual encoders kept frozen throughout the experiments. Additionally, to enhance the clinical relevance of our findings, we conduct experiments on a state-of-the-art medical-domain MLLM, \texttt{HuatuoGPT}, to validate the generalizability and robustness of editing methods under realistic medical scenarios.
\\To systematically assess the performance of knowledge editing methods under different levels of knowledge injection, we adopt two complementary evaluation settings. \textbf{Single Editing} focuses on the precise injection of an individual medical fact and evaluates the model's immediate response accuracy on associated queries immediately after the update, effectively reflecting the model’s responsiveness to targeted knowledge modifications. \textbf{Lifelong Editing}, on the other hand, simulates continuous and incremental knowledge evolution in real-world applications by sequentially injecting 50, 250, 500, 750, and 1000 knowledge entries. This setup is used to examine the model’s ability to retain newly added knowledge, mitigate cumulative interference, and robustly resist catastrophic forgetting during long-term editing.

\begin{figure*}[!ht]
    \centering
    \includegraphics[width=\linewidth]{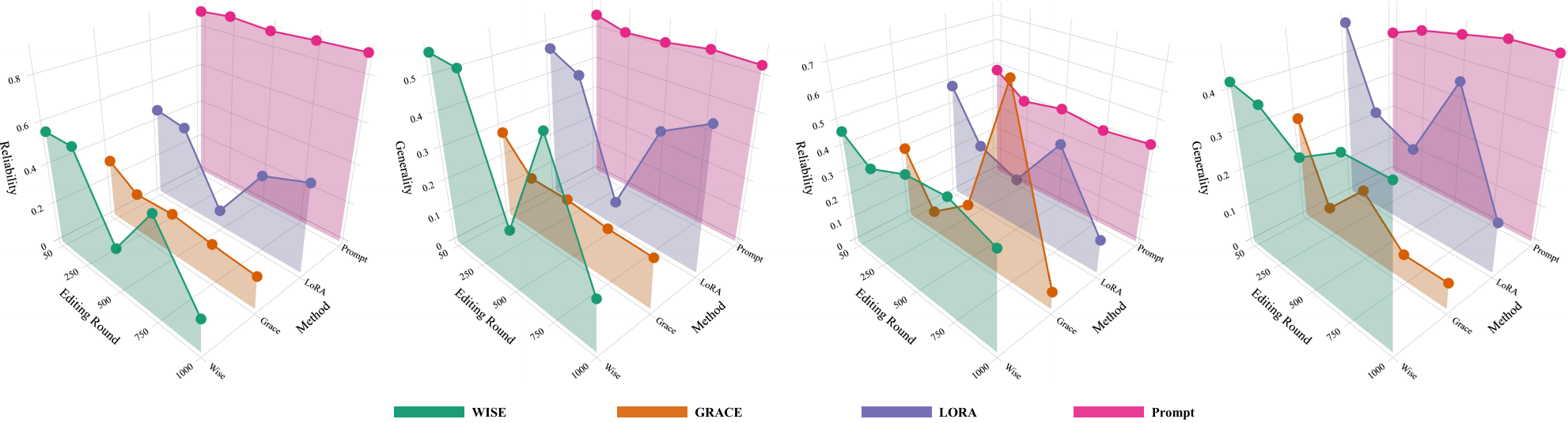}
    \caption{
        Lifelong editing results of Qwen2vl-7B using four methods: Promp, WISE, GRACE, and LoRA. Left two subplots correspond to \textit{understanding} tasks, and right two subplots correspond to \textit{reasoning} tasks.
    }
    \label{fig:LifeLongEditResult2.jpg}
\end{figure*}

\subsection{Main Results}

\subsubsection{For Q1: How do current knowledge-editing methods perform overall in multimodal medical scenarios?}
Experimental results in multimodal medical scenarios (Table~\ref{tab:table2}) reveal significant disparities in performance across knowledge-editing methods, driven by inherent trade-offs rooted in their design principles. Broadly, methods fall into two categories: external guidance approaches like Prompt, which injects knowledge via in-context learning without altering model parameters, and internal parametric methods such as WISE, LoRA, and GRACE. Prompt effectively leverages the model’s language priors to achieve high Reliability and Generality, but their lack of structural constraint leads to poor Locality, causing interference with unrelated tasks. In contrast, parametric methods introduce targeted structural updates that ensure near-perfect Locality by isolating the edit’s impact. However, their efficacy is inconsistent and generalization is limited. LoRA, which uses low-rank adapters, performs well on some models (e.g., 0.9033 on \texttt{HuaTuoGPT-7B}) but collapses on others (e.g., 0.1852 on \texttt{QWen2-VL}), exposing its dependency on model architecture. WISE relies on external memory, and its performance is tightly coupled with how well the memory module integrates into the model’s reasoning pathways. GRACE, which injects new parameters, achieves strong reliability but suffers from poor generality (e.g., 0.4295), indicating brittle, over-localized edits that resemble memorization rather than knowledge integration. In essence, current methods face a structural dilemma: they either promote generalizable knowledge at the expense of control, or enforce localized precision while failing to support transferable reasoning.

\subsubsection{For Q2: How does editing effectiveness differ between the Understanding and Reasoning tasks?}
A comparison between \textit{Understanding} and \textit{Reasoning} tasks (Table~\ref{tab:table2}) reveals not a simple complexity gap, but fundamentally a mechanism-driven interaction between editing methods and task types. Methods like WISE and LoRA, which rely on internal parameter updates (e.g., memory injection or low-rank tuning), are effective for localized factual edits but struggle to generalize across temporally and multimodally complex reasoning inputs due to their scoped, non-generative nature. However, this is not universally true. GRACE, for instance, achieves near-perfect Reliability on Reasoning in \texttt{QWen2-VL} but underperforms on Understanding. Its injected parameters may act as shortcuts in the inference chain, enabling multi-step reasoning but failing to generalize due to low semantic abstraction. Similarly, Prompt performs better on Reasoning in \texttt{HuaTuoGPT-7B}, consistent with their design:Pprompt operates at the input level, and when the base model has strong inferential capabilities, context injection can more effectively propagate through reasoning paths than in factual recall. Ultimately, editing success hinges less on task difficulty and more on alignment between editing scope, cognitive demands, and model architecture. \textit{Understanding} tasks require stable factual rewrites across variants, while Reasoning tasks demand edits that integrate effectively into dynamic inference mechanisms.

\begin{figure}[!ht]
    \centering
    \begin{subfigure}[b]{0.48\textwidth}      
        \centering
        \includegraphics[scale=0.28]{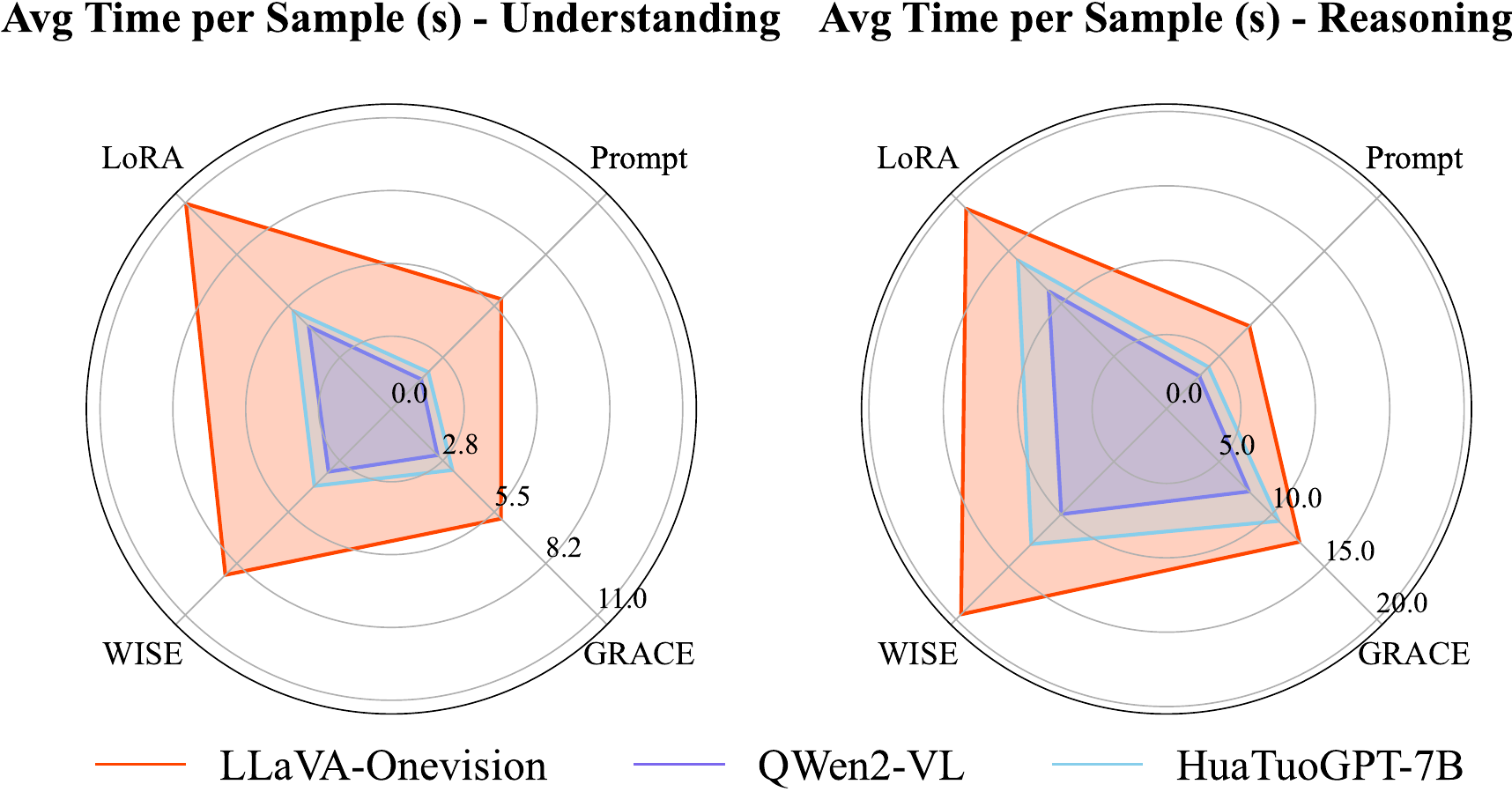}
        \caption{Average time (s) per sample for different editing methods on \textit{Understanding} and \textit{Reasoning} tasks.}
        \label{fig:EfficiencyAnalysis-time}
    \end{subfigure}
    \hfill
    \begin{subfigure}[b]{0.48\textwidth}
        \centering
        \includegraphics[scale=0.34]{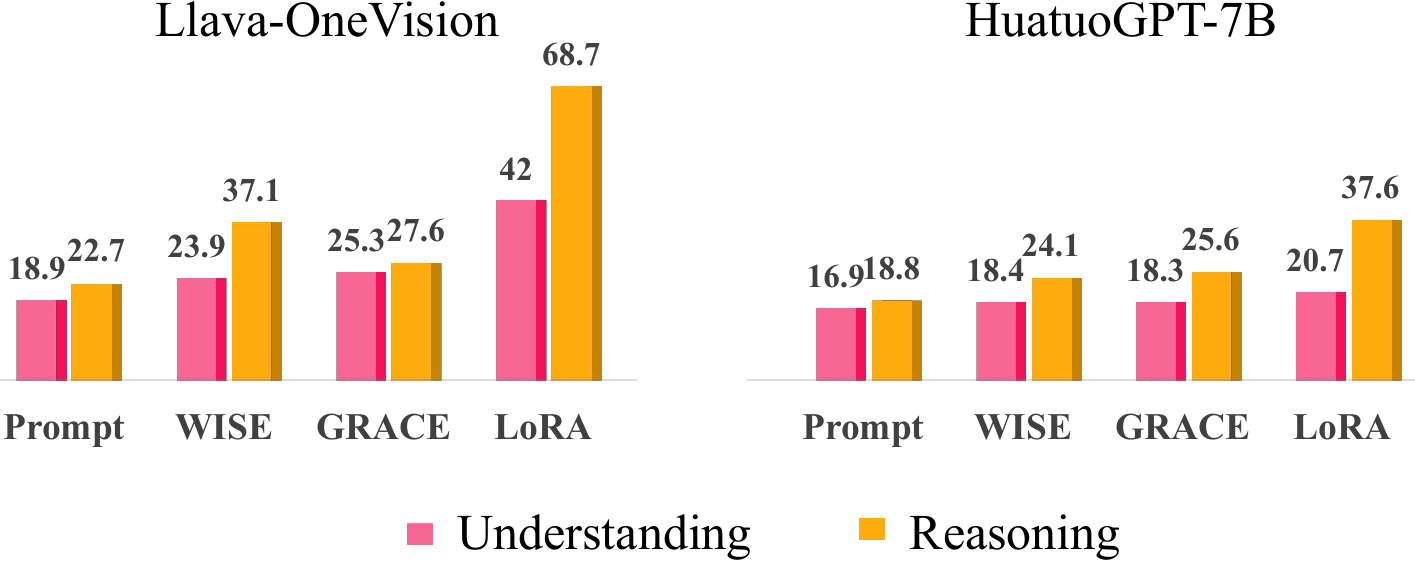}
        \caption{Peak GPU memory (GB) per sample for different editing methods on \textit{Understanding} and \textit{Reasoning} tasks.}
        \label{fig:EfficiencyAnalysis-memory}
    \end{subfigure}
    \caption{Efficiency comparison of different editing methods using LLaVA-OneVision and HuaTuoGPT-7B.}
    \label{fig:EfficiencyAnalysis}
\end{figure}

\begin{figure*}[!htp]
    \centering
   \includegraphics[scale=0.25]{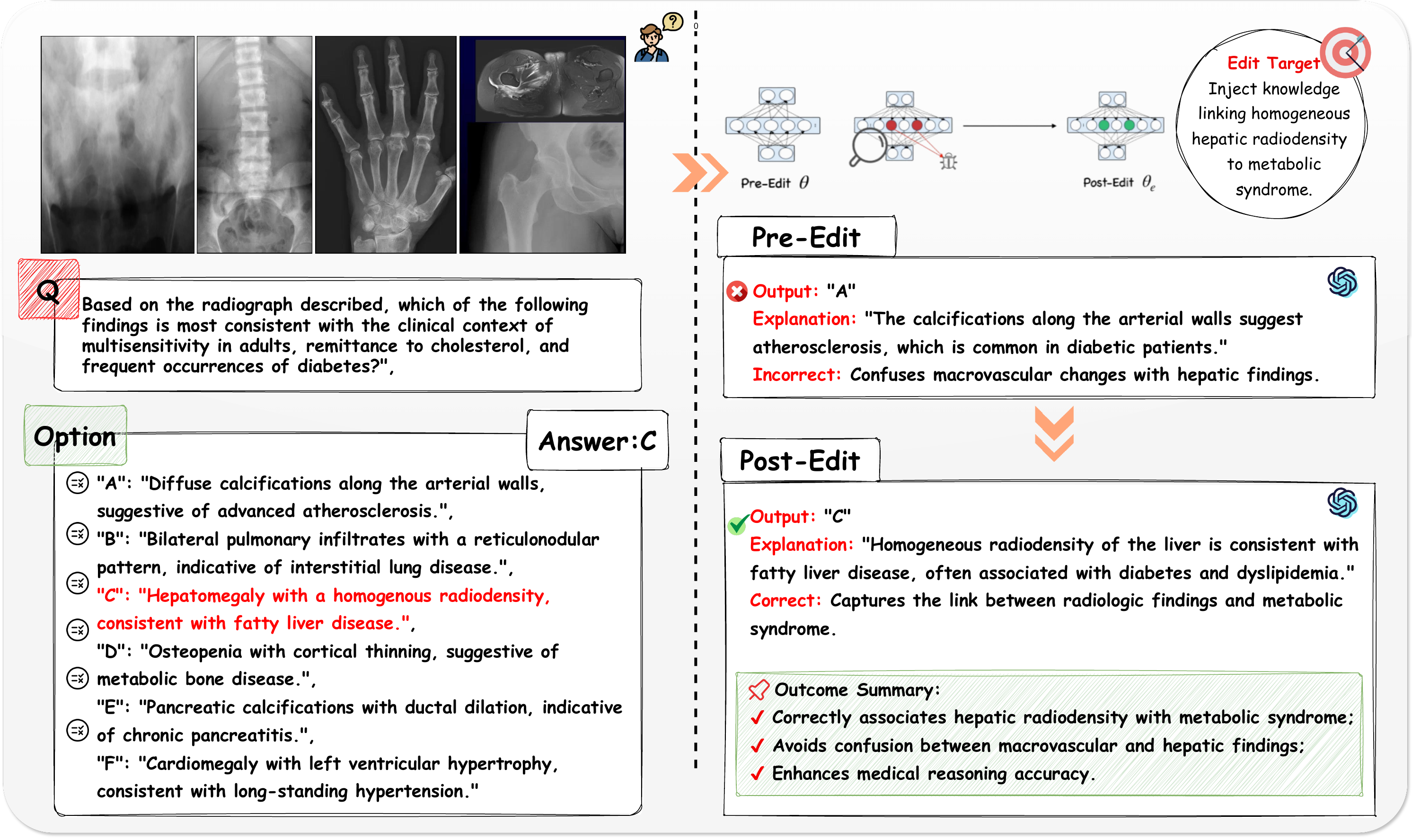}
    \caption{Case analysis of editing LLaVA-OneVision with WISE.}
    \label{fig:CaseAnalysis}
\end{figure*}

\subsubsection{For Q3: How does edit scale (Single Knowledge Editing vs. Lifelong Knowledge Editing) affect model performance and stability?}
Our investigation into continuous knowledge editing reveals that the primary challenge is not systematic performance degradation, but significant performance volatility and method-dependent stability as the number of edits increases, as shown in Figure~\ref{fig:LifeLongEditResult2.jpg}. Regarding Locality, weight-modifying methods like WISE, GRACE, and LoRA consistently achieve near-perfect T-Locality and M-Locality scores (1.0), demonstrating a strong and consistent ability to prevent interference with unrelated knowledge, which is why these metrics were omitted from the visual plots for clarity. In sharp contrast, the prompt-based method shows persistently lower locality scores (around 0.75-0.80), indicating chronic and compounding side effects. However, a critical trade-off emerges in Reliability and Generality. The prompt-based method offers remarkable stability with constant, predictable performance. Conversely, WISE, GRACE, and LoRA suffer from extreme instability, with erratic performance featuring unpredictable peaks and troughs. This is starkly exemplified by GRACE's Reliability score, which spiked at 750 edits before collapsing. This highlights a fundamental dilemma: prompt-based methods provide stable but leaky edits with poor locality, whereas current weight-space methods offer well-contained but volatile edits whose success becomes chaotic and unreliable over time.




\subsection{Efficiency Comparison}
\paragraph{Time Consumption Analysis.}
As shown in Figure~\ref{fig:EfficiencyAnalysis-time}, the variation in editing time directly reflects the structural complexity and computational overhead of different methods. The high latency observed in LoRA and WISE stems from their deep interventions in model parameters—LoRA introduces low-rank adapters, while WISE invokes external memory and re-encodes semantics. These intrusive operations significantly extend the computational path, especially in multimodal reasoning tasks. In contrast, Prompt performs localized and lightweight edits, relying solely on direct manipulation within the activation space, without reconstructing full computational graphs, thereby achieving faster response. GRACE, as a parameter-injection method, adopts a more modular structural design, resulting in intermediate latency. Overall, the more intrusive the editing mechanism, the harder it becomes to support low-latency updates required in time-sensitive scenarios such as clinical applications.

\paragraph{Memory Consumption Analysis.}
As illustrated in Figure~\ref{fig:EfficiencyAnalysis-memory}, memory usage intuitively reflects differences in intermediate state storage and structural overhead. Although LoRA is considered parameter-efficient, in practice it incurs the highest memory footprint due to the need to maintain adapter weights and gradient caches, especially when handling complex, high-dimensional inputs. WISE and GRACE also show significant increases in memory consumption in multi-frame reasoning tasks, largely attributed to the inclusion of external modules or additional computational paths. In contrast, Prompt demonstrates the most stable memory profile, benefiting from its local editing strategy that avoids loading additional structures or states, thus exhibiting stronger resource adaptability. This makes Prompt more suitable for deployment in highly resource-constrained edge devices or real-world medical systems. In general, memory efficiency not only affects scalability but also reflects the overall resource-friendliness of the method’s architectural design. Ultimately, these findings position the Prompt method as the most practical choice for scalable deployment.

\subsection{Case Analysis}
As illustrated in Figure~\ref{fig:CaseAnalysis}, applying the WISE method to \texttt{LLaVA-OneVision} successfully injects the missing association between homogeneous hepatic radiodensity and metabolic syndrome. Prior to editing, the model incorrectly selects a macrovascular diagnosis (Option A), clearly reflecting a failure to distinguish vascular calcifications from hepatic findings. After the targeted edit, the model correctly selects Option C, demonstrating an understanding that hepatomegaly with homogeneous radiodensity is indicative of fatty liver disease—a manifestation commonly linked to various metabolic disorders. This shift clearly reflects precise localization of the injected knowledge and improved multimodal reasoning, despite the edit being applied only to the language module. The overall outcome highlights the cross-modal coupling in vision-language models: linguistic edits can effectively propagate to correct grounded diagnostic reasoning without fine-tuning the vision backbone.

\section{Conclusion and Future Outlook}
In this paper, we present \textbf{MultiMedEdit}, the first comprehensive benchmark for evaluating knowledge editing in multimodal medical scenarios. Through systematic evaluations, we reveal that current editing methods exhibit significant limitations in accuracy, order sensitivity, and catastrophic forgetting, particularly in complex tasks requiring deep semantic understanding and robust clinical reasoning. While our benchmark provides a solid foundation, it is currently limited by a primary focus on question-answering tasks and lacks deeper insights into the interpretability of editing mechanisms. Future work will further expand task diversity and complexity, thoroughly analyze model-internal representations to enhance interpretability, and develop novel domain-specific, minimally invasive editing methods for the sustainable long-term evolution of medical knowledge.

\bibliography{aaai2026}

\makeatletter
\@ifundefined{isChecklistMainFile}{
  \newif\ifreproStandalone
  \reproStandalonetrue
}{
  \newif\ifreproStandalone
  \reproStandalonefalse
}
\makeatother

\setlength{\pdfpagewidth}{8.5in}
\setlength{\pdfpageheight}{11in}
\frenchspacing

\setlength{\leftmargini}{20pt}
\makeatletter
\def\@listi{\leftmargin\leftmargini \topsep .5em \parsep .5em \itemsep .5em}
\def\@listii{\leftmargin\leftmarginii \labelwidth\leftmarginii \advance\labelwidth-\labelsep \topsep .4em \parsep .4em \itemsep .4em}
\def\@listiii{\leftmargin\leftmarginiii \labelwidth\leftmarginiii \advance\labelwidth-\labelsep \topsep .4em \parsep .4em \itemsep .4em}
\makeatother

\setcounter{secnumdepth}{0}
\renewcommand\thesubsection{\arabic{subsection}}
\renewcommand\labelenumi{\thesubsection.\arabic{enumi}}

\newcounter{checksubsection}
\newcounter{checkitem}[checksubsection]

\newcommand{\checksubsection}[1]{%
  \refstepcounter{checksubsection}%
  \paragraph{\arabic{checksubsection}. #1}%
  \setcounter{checkitem}{0}%
}

\newcommand{\checkitem}{%
  \refstepcounter{checkitem}%
  \item[\arabic{checksubsection}.\arabic{checkitem}.]%
}
\newcommand{\question}[2]{\normalcolor\checkitem #1 #2 \color{blue}}
\newcommand{\ifyespoints}[1]{\makebox[0pt][l]{\hspace{-15pt}\normalcolor #1}}

\ifreproStandalone
\begin{document}
\fi

\section*{Reproducibility Checklist}









\checksubsection{General Paper Structure}
\begin{itemize}

\question{Includes a conceptual outline and/or pseudocode description of AI methods introduced}{(yes/partial/no/NA)}
yes

\question{Clearly delineates statements that are opinions, hypothesis, and speculation from objective facts and results}{(yes/no)}
no

\question{Provides well-marked pedagogical references for less-familiar readers to gain background necessary to replicate the paper}{(yes/no)}
yes

\end{itemize}
\checksubsection{Theoretical Contributions}
\begin{itemize}

\question{Does this paper make theoretical contributions?}{(yes/no)}
no

\ifyespoints{\vspace{1.2em}If yes, please address the following points:}
\begin{itemize}

\question{All assumptions and restrictions are stated clearly and formally}{(yes/partial/no)}
NA

\question{All novel claims are stated formally (e.g., in theorem statements)}{(yes/partial/no)}
NA

\question{Proofs of all novel claims are included}{(yes/partial/no)}
NA

\question{Proof sketches or intuitions are given for complex and/or novel results}{(yes/partial/no)}
NA

\question{Appropriate citations to theoretical tools used are given}{(yes/partial/no)}
NA

\question{All theoretical claims are demonstrated empirically to hold}{(yes/partial/no/NA)}
NA

\question{All experimental code used to eliminate or disprove claims is included}{(yes/no/NA)}
NA

\end{itemize}
\end{itemize}

\checksubsection{Dataset Usage}
\begin{itemize}

\question{Does this paper rely on one or more datasets?}{(yes/no)}
yes

\ifyespoints{If yes, please address the following points:}
\begin{itemize}

\question{A motivation is given for why the experiments are conducted on the selected datasets}{(yes/partial/no/NA)}
yes

\question{All novel datasets introduced in this paper are included in a data appendix}{(yes/partial/no/NA)}
partial

\question{All novel datasets introduced in this paper will be made publicly available upon publication of the paper with a license that allows free usage for research purposes}{(yes/partial/no/NA)}
yes

\question{All datasets drawn from the existing literature (potentially including authors' own previously published work) are accompanied by appropriate citations}{(yes/no/NA)}
yes

\question{All datasets drawn from the existing literature (potentially including authors' own previously published work) are publicly available}{(yes/partial/no/NA)}
yes

\question{All datasets that are not publicly available are described in detail, with explanation why publicly available alternatives are not scientifically satisficing}{(yes/partial/no/NA)}
NA

\end{itemize}
\end{itemize}

\checksubsection{Computational Experiments}
\begin{itemize}

\question{Does this paper include computational experiments?}{(yes/no)}
yes

\ifyespoints{If yes, please address the following points:}
\begin{itemize}

\question{This paper states the number and range of values tried per (hyper-) parameter during development of the paper, along with the criterion used for selecting the final parameter setting}{(yes/partial/no/NA)}
partial

\question{Any code required for pre-processing data is included in the appendix}{(yes/partial/no)}
yes

\question{All source code required for conducting and analyzing the experiments is included in a code appendix}{(yes/partial/no)}
yes

\question{All source code required for conducting and analyzing the experiments will be made publicly available upon publication of the paper with a license that allows free usage for research purposes}{(yes/partial/no)}
yes

\question{All source code implementing new methods have comments detailing the implementation, with references to the paper where each step comes from}{(yes/partial/no)}
partial

\question{If an algorithm depends on randomness, then the method used for setting seeds is described in a way sufficient to allow replication of results}{(yes/partial/no/NA)}
yes

\question{This paper specifies the computing infrastructure used for running experiments (hardware and software), including GPU/CPU models; amount of memory; operating system; names and versions of relevant software libraries and frameworks}{(yes/partial/no)}
partial

\question{This paper formally describes evaluation metrics used and explains the motivation for choosing these metrics}{(yes/partial/no)}
yes

\question{This paper states the number of algorithm runs used to compute each reported result}{(yes/no)}
yes

\question{Analysis of experiments goes beyond single-dimensional summaries of performance (e.g., average; median) to include measures of variation, confidence, or other distributional information}{(yes/no)}
partial

\question{The significance of any improvement or decrease in performance is judged using appropriate statistical tests (e.g., Wilcoxon signed-rank)}{(yes/partial/no)}
yes

\question{This paper lists all final (hyper-)parameters used for each model/algorithm in the paper’s experiments}{(yes/partial/no/NA)}
partial

\end{itemize}
\end{itemize}

\ifreproStandalone
\end{document}
\fi

\end{document}


\maketitle


\appendix
\section{Appendix A: Medical VQA Datasets}   

\subsection{A.1 Medical Multimodal QA Datasets}

\textbf{MedFrameQA} is a multi-frame visual question answering benchmark designed for clinical reasoning. It includes temporally ordered CT slices, ultrasound videos, and associated clinical questions targeting lesion evolution, treatment evaluation, and temporal decision-making. Each question is paired with multiple images from real clinical workflows, emphasizing the model’s capacity for temporal fusion and cross-frame inference. In this work, we select a high-quality subset for constructing reasoning-based knowledge editing tasks.

\noindent\textbf{PMC-VQA} is built on figures and captions from biomedical articles in PubMed Central. Each instance consists of a static medical image (e.g., radiology or pathology) and a corresponding textual context question. The dataset evaluates a model's ability to jointly interpret biomedical visuals and domain-specific descriptions. We utilize a filtered subset from PMC-VQA to construct the understanding-focused portion of our benchmark.

\noindent\textbf{MedXpertQA} provides expert-level clinical QA pairs covering complex reasoning and professional medical logic. The dataset spans multiple specialties and includes step-wise explanations and domain-aligned answer rationales. We incorporate samples from MedXpertQA to increase task diversity and enhance coverage of complex medical knowledge in our editing benchmark.

\subsection{A.2 Locality Evaluation Datasets}
\textbf{Natural Questions (NQ)} is a large-scale QA dataset collected from anonymized Google search queries. Each question is grounded in factual or commonsense knowledge, with answers derived from Wikipedia. We randomly sample a subset of NQ to evaluate \textit{textual locality}—i.e., whether knowledge editing in the medical domain disrupts model behavior on unrelated general-domain tasks.

\noindent\textbf{VQAv2} is a widely used benchmark for visual question answering over natural images. It contains open-domain visual questions involving attributes, counting, spatial reasoning, and commonsense judgments. As it is structurally similar to medical VQA but lacks domain-specific knowledge, we use it to assess \textit{multimodal locality}—the robustness of visual reasoning in non-medical contexts post-editing.

\section{Appendix B: Evaluation Metrics}

\subsection{Reliability}
Reliability measures the accuracy of the post-edit model $f_{\theta_e}$ in producing the intended output for the edited input $(x_e, y_e)$. It is computed as the expected success rate over the edit set:
\begin{equation}
M_{\text{rel}} = \mathbb{E}_{(x_e, y_e) \sim \mathcal{D}_e} \left[ f_{\theta_e}(x_e) = y_e \right]
\end{equation}
\textbf{Example:} If the model is updated to answer the question “What is the treatment for NAFLD?” with “Lifestyle intervention” instead of “Aspirin,” this metric checks whether $f_{\theta_e}$ correctly returns the new answer.

\subsection{Generality}
Generality reflects whether the edited model can correctly respond to semantically equivalent but rephrased or modality-varied inputs. It is evaluated in both textual and multimodal forms:
\begin{equation}
M_{\text{gen}}^{\text{text}} = \mathbb{E}_{x_r \sim \mathcal{N}(x_e)} \left[ f_{\theta_e}(x_r) = y_e \right]
\end{equation}
\begin{equation}
M_{\text{gen}}^{\text{mm}} = \mathbb{E}_{m_r \sim \mathcal{N}(m_e)} \left[ f_{\theta_e}(x_e, m_r) = y_e \right]
\end{equation}
Here, $\mathcal{N}(x)$ denotes a neighborhood of semantically equivalent variants.

\noindent\textbf{Example:} Given the original query “Which imaging finding indicates fatty liver?”, this metric evaluates whether variations like “What radiologic evidence supports hepatic steatosis?” produce the correct edited output.

\subsection{Locality}
Locality assesses the degree to which unrelated outputs remain stable after editing. It captures potential side effects introduced by the edit. The textual and multimodal locality metrics are computed as:
\begin{equation}
M_{\text{loc}}^{\text{text}} = \mathbb{E}_{(x, y) \sim \mathcal{D}_{\text{loc}}} \left[ f_{\theta_e}(x) = f_\theta(x) \right]
\end{equation}
\begin{equation}
M_{\text{loc}}^{\text{mm}} = \mathbb{E}_{(x, m, y) \sim \mathcal{D}_{\text{loc-v}}} \left[ f_{\theta_e}(x, m) = f_\theta(x, m) \right]
\end{equation}

\noindent\textbf{Example:} If an edit targets liver disease reasoning, locality verifies that predictions for unrelated topics such as orthopedic injuries or cardiovascular assessment remain unchanged.

\begin{figure*}[!ht]
    \centering
   \includegraphics[width=\linewidth]{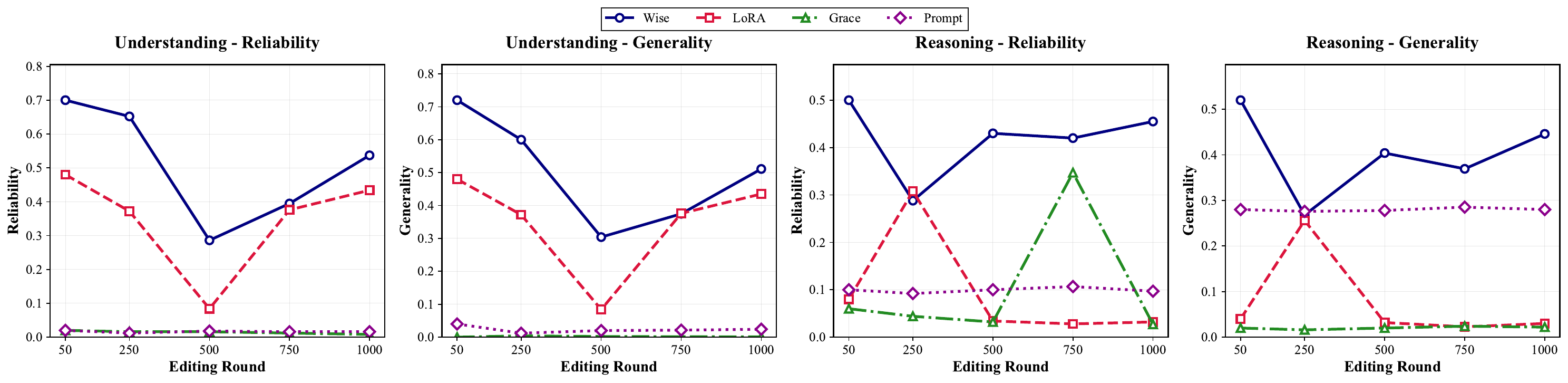}
   \caption{Lifelong editing performance of HuatuoGPT-7B using different editing approaches (Wise, LoRA, Grace, Prompt). The figure reports Reliability and Generality scores on both \textit{understanding} and \textit{reasoning} tasks under varying numbers of edits.}

    \label{fig:LifeLong}
\end{figure*}

\section{Appendix C: Lifelong Edit on HuatuoGPT-7B}
The Lifelong Edit Result on HuatuoGPT-7B is shown in Figure~\ref{fig:LifeLong}. For all methods across both understanding and reasoning tasks, T-Locality and M-Locality remain consistently high (mostly above 0.8 or even 1.0), indicating that the edits are well-localized both in terms of task and memory, with minimal unintended interference.

\section{Appendix D: Benchmark Construction Details}
\begin{figure*}[htp]
    \centering
   \includegraphics[scale=0.38]{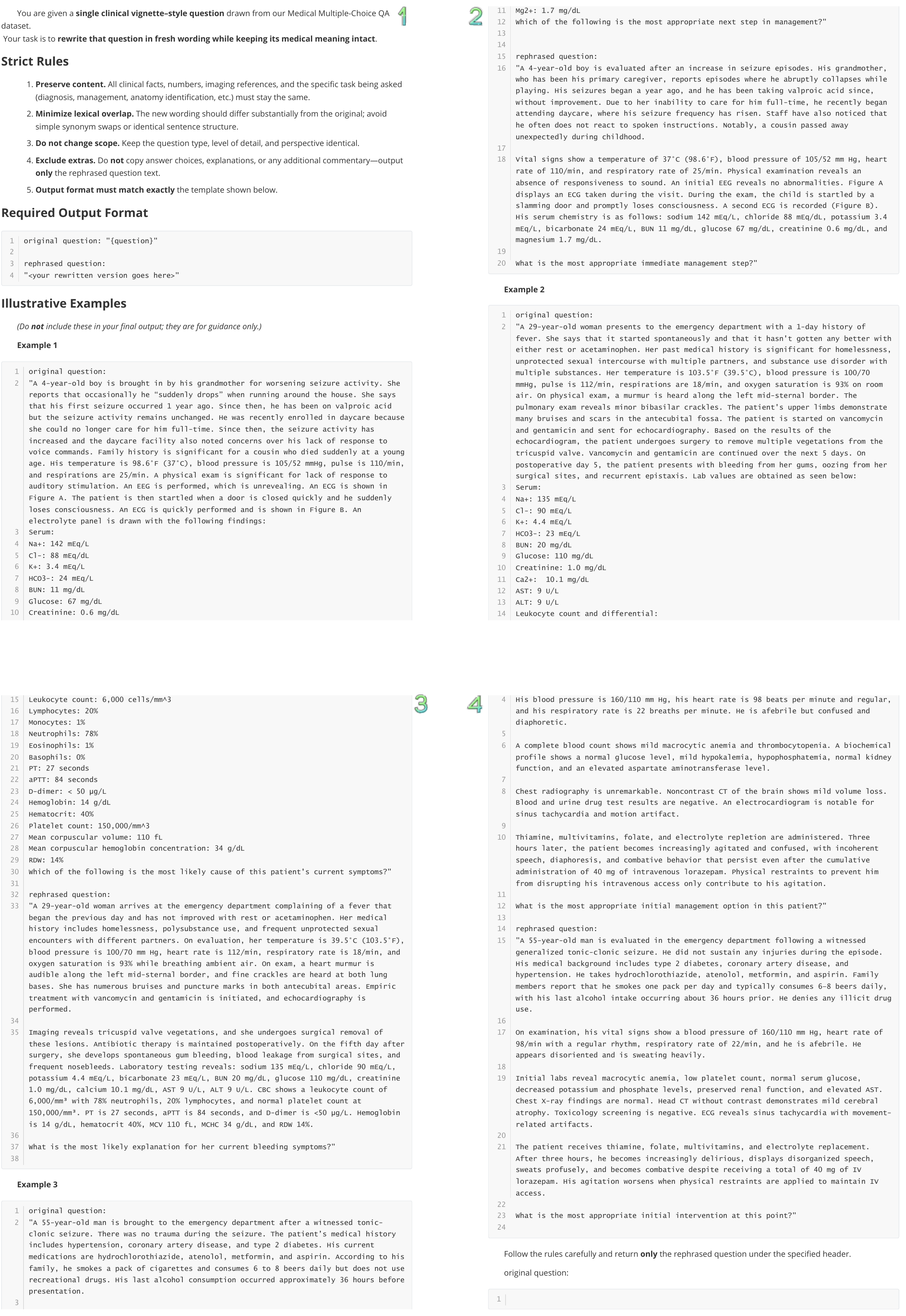}
    \caption{Prompt template of question rephrase.}
    \label{fig:PromptRephrase}
\end{figure*}

The prompt templates are as shown in Figure \ref{fig:PromptRephrase}.

\section{Appendix E: Samples from MultiMedBench}
\begin{figure*}[htp]
    \centering
   \includegraphics[scale=0.7]{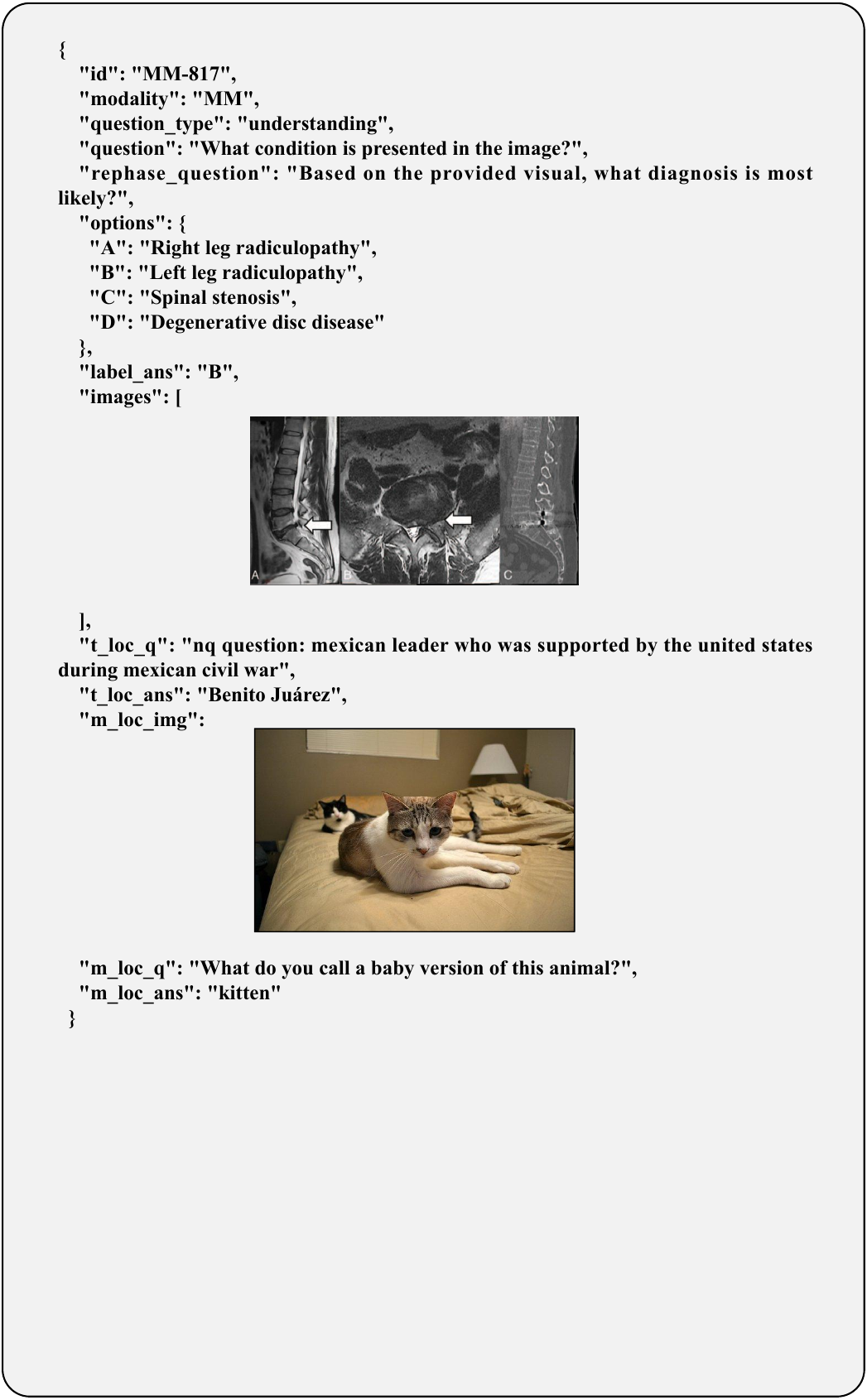}
    \caption{Prompt template of question rephrase.}
    \label{fig:CaseDataUnderstanding}
\end{figure*}

\begin{figure*}[htp]
    \centering
   \includegraphics[scale=0.7]{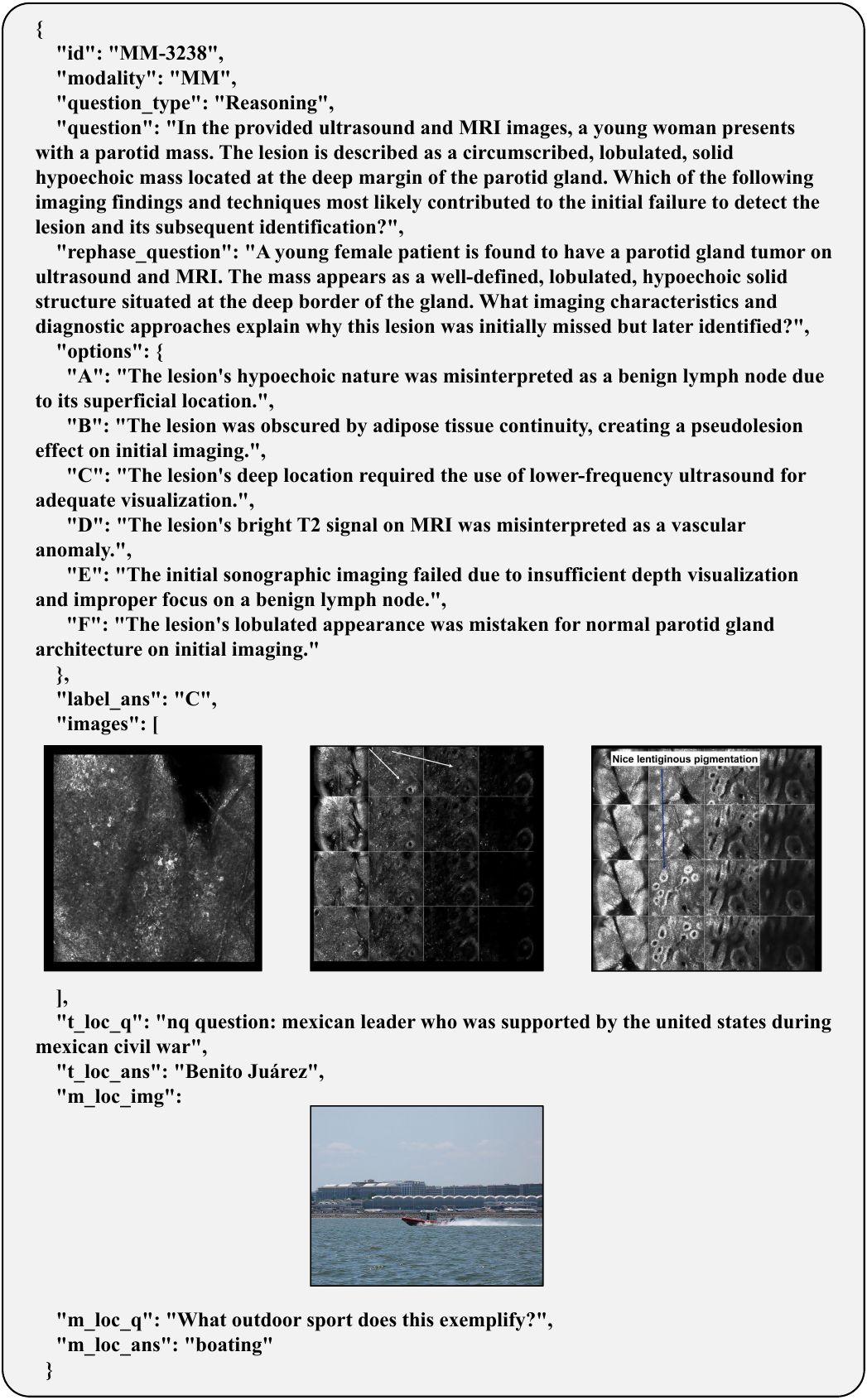}
    \caption{Prompt template of question rephrase.}
    \label{fig:CaseDataReasoning}
\end{figure*}

Representative cases from MultiMedBench are illustrated in Figure~\ref{fig:CaseDataUnderstanding} and Figure~\ref{fig:CaseDataReasoning}, covering both understanding and reasoning tasks. We have selected 50 examples for each category, which are included in the accompanying ZIP file.

